\pdfoutput=1

\documentclass[11pt]{article}

\usepackage[]{acl}

\usepackage{times}
\usepackage{latexsym}
\usepackage{enumitem}
\usepackage{graphicx}
\usepackage{bbm}

\usepackage[T1]{fontenc}

\usepackage[utf8]{inputenc}

\usepackage{microtype}
\usepackage{xcolor}

\definecolor{redd}{HTML}{B85450}
\definecolor{greeen}{HTML}{82B366}

\newcommand{\citeay}[1]{\citeauthor{#1}., \citeyear{#1}}
\newcommand{\mpp}{\textsc{mpp}}

\title{Language model acceptability judgements are not always robust to context}

\author{Koustuv Sinha $^{*,\infty}$ \quad Jon Gauthier $^{*,1}$ \\ {\bf Aaron Mueller} $^{\dagger, 3}$ \quad {\bf Kanishka  Misra} $^{\dagger, 3}$ \quad {\bf Keren Fuentes} $^\infty$ \\ {\bf Roger Levy} $^1$ \quad {\bf Adina Williams} $^\infty$ \\
$^\infty$ Meta AI; $^1$ MIT $^2$ Purdue University $^3$ Johns Hopkins \\
$^*,\dagger$ Equal contributions \\
\texttt{koustuvs@meta.com, jon@gauthiers.net}}

\begin{document}
\maketitle
\begin{abstract}

Targeted syntactic evaluations of language models ask whether models show stable preferences for syntactically acceptable content over minimal-pair unacceptable inputs. Most targeted syntactic evaluation datasets ask models to make these judgements with just a single context-free sentence as input. This does not match language models' training regime, in which input sentences are always highly contextualized by the surrounding corpus.
This mismatch raises an important question: how robust are models' syntactic judgements in different contexts?
In this paper, we investigate the stability of language models' performance on targeted syntactic evaluations as we vary properties of the input context: the length of the context, the types of syntactic phenomena it contains, and whether or not there are violations of grammaticality. 
We find that model judgements are generally robust when placed in randomly sampled linguistic contexts. However, they are substantially unstable for contexts containing syntactic structures matching those in the critical test content. Among all tested models (GPT-2 and five variants of OPT), we significantly improve models' judgements by providing contexts with matching syntactic structures, and conversely significantly worsen them using unacceptable contexts with matching but violated syntactic structures. This effect is amplified by the length of the context, except for unrelated inputs. 
We show that these changes in model performance are not explainable by simple features matching the context and the test inputs, such as lexical overlap and dependency overlap. This sensitivity to highly specific syntactic features of the context can only be explained by the models' implicit in-context learning abilities.

\end{abstract}

\section{Introduction}


\begin{figure}[t]
    \centering
    \includegraphics[width=\columnwidth]{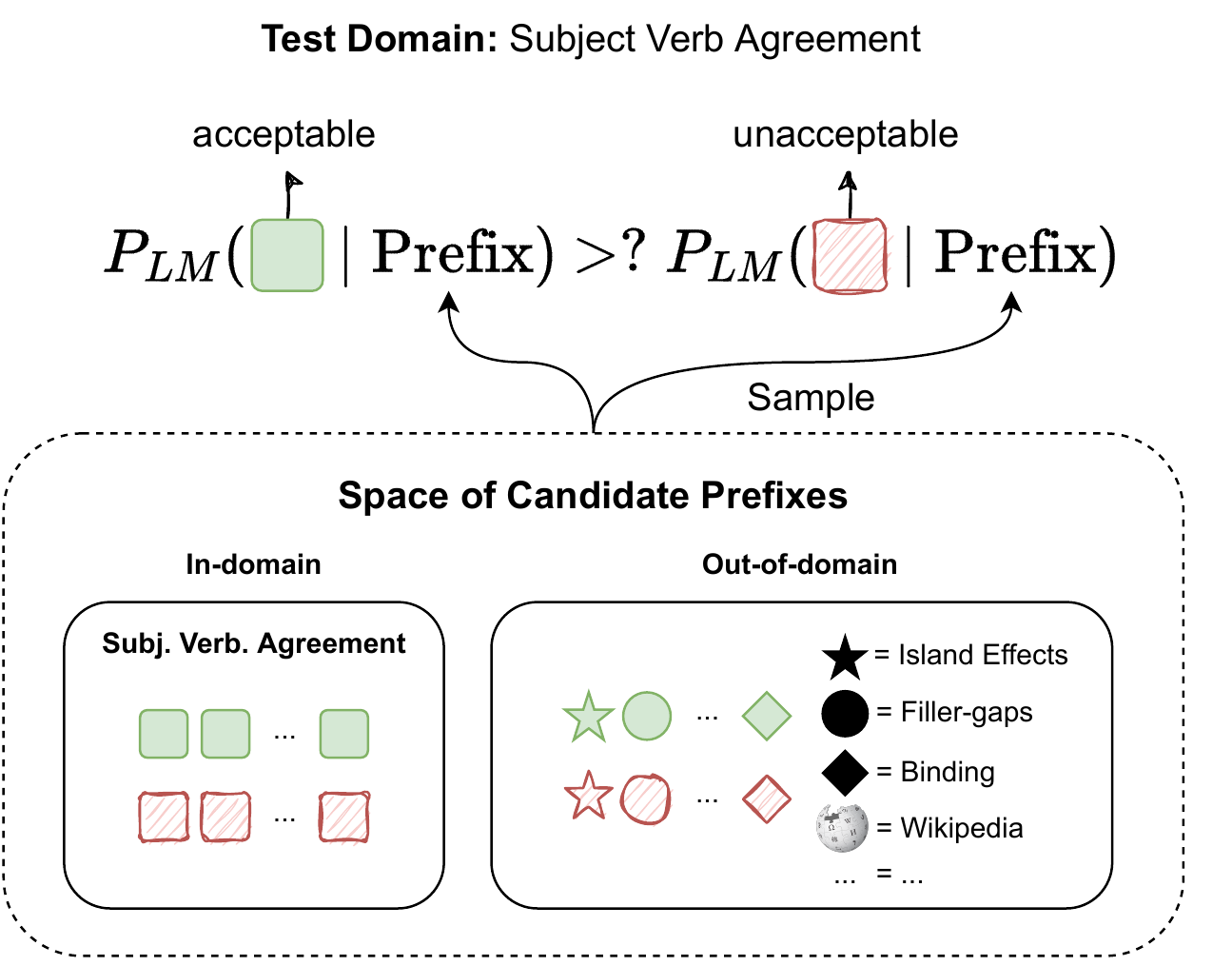}
    \caption{We measure the impact of different contexts on the performance of an LM on linguistic acceptability tasks by prefixing sentences (here, sourced from subject-verb agreement challenge sets) from a diverse collection of sources. Each block represents a sentence. \textcolor{redd}{\textbf{Red}} striped blocks are unacceptable sentences within a given task, while \textcolor{greeen}{\textbf{green}} solid ones are acceptable. We also vary the diversity of prefixes by sampling them from tasks/datasets different from the test domain (indicated by shape).}
    \label{fig:my_label}
\end{figure}

The unprecedented progress in the development of neural large language models \citep[LLMs;][]{devlin-etal-2019-bert, radford2019language, brown2020language, zhang2022opt} has been accompanied by a comparable proliferation of methods that aim to better understand and characterize models' linguistic capacities \citep[][\textit{i.a.}]{linzen2016assessing, ettinger-etal-2016-probing, alishahi2019analyzing, hu-etal-2020-systematic, jeretic-etal-2020-natural, mueller-etal-2020-cross}. Of the many methods for this, the minimal-pair paradigm (\mpp{}), which is methodologically standard in linguistics, has emerged as a popular approach to evaluate models' knowledge of linguistic phenomena in an unsupervised manner \citep{marvin-linzen-2018-targeted, kann-etal-2019-verb, warstadt-etal-2019-neural, warstadt-etal-2020-blimp, misra2022comps}. 
Under the \mpp{}, models are presented with datasets containing pairs of minimally differing text sequences---usually differing in word order or in a few tokens---one of which is deemed by humans to be acceptable and the other unacceptable. Drawing on the LLMs' trained ability to produce probabilities over token sequences, we can evaluate them according to the \mpp{} by testing whether models assign relatively greater probability to the acceptable sequence. 

Studies that employ \mpp{} datasets generally compare the probability of two stand-alone text sequences without any explicit linguistic context (or, the probability of two words that are part of some stand-alone sentence). However, this is not a naturalistic or realistic approach: utterances usually occur \emph{in some linguistic context}, where the context itself could affect linguistic preferences. The syntactic priming literature investigates the effect of linguistic contexts to some extent, but mostly in a constrained setting with only one or a small number of context sentences \cite{van-schijndel-linzen-2018-neural, prasad-etal-2019-using}.
The interaction of context with minimal pair accuracies remains underexplored for multi-sentence contexts, despite the fact that multi-sentence inputs are more likely for many NLP tasks---especially with the rise of prompting and in-context learning \citep{brown2020language, schick-schutze-2021-just}. Furthermore, Transformer-based language models are typically trained on large sequences, where masked tokens are predicted given a completely full context window, consisting of many sentences. It is unclear how to evaluate \mpp\ by utilizing this context window, given recent research that has raised questions about the sentence representations acquired in long-form input \cite{sinha2022curious}.

We evaluate the sensitivity of LLMs' acceptability preferences in a more realistic evaluation setting, with one or more additional sentences in the input context. We focus on LLM sensitivity to three particular features of the context: (1) the length of the input sequence, (2) the similarity of the context to the minimal pair being judged, and (3) whether the prefix context contains ungrammatical language.
\autoref{fig:my_label} illustrates our method at a high level: For a given \mpp{} dataset, we generate new, minimal pair test examples for a given syntactic phenomenon by artificially simulating a long context window. Specifically, we prepend the given test example pair with context constructed from sentences drawn from Wikipedia (\emph{unrelated context}), and compare it with contexts constructed with minimal-pair sentences from  the same (\emph{in-domain}) or different (\emph{out-of-domain}) syntactic phenomena in the \mpp{} dataset.

We find that the model's judgements are highly robust to the presence of unrelated, out-of-domain Wikipedia sentences in the context, regardless of the size of the context.
However, we observe strong sensitivity to in-domain context manipulations. As the context length increases, acceptable, grammatical in-domain contexts improve the models' judgements significantly. Conversely, we observe a strong negative effect of exposing the model to longer and longer ungrammatical or unacceptable context: models' judgements degrade drastically, performing far below chance. 
This sensitivity is specific to the particular type of syntactic structural similarity of the context: we do not see the same degree of improvement/degradation in prediction behavior for contexts consisting of out-of-domain sentences of valid or violated syntactic structures.

To better understand our results, we performed several exploratory analyses. 
We explored several linguistic features (lexical overlap, dependency overlap) to explain whether syntactic similarity can explain our results (\S\ref{sec:similarity_analysis}) and found that the trends cannot be explained only by these low-level overlap features. 
We also investigated model calibrations when subjected to prefixed stimuli using perplexity margins, to explain the changes in accuracy with different types of prefixes (Appendix \S\ref{sec:margin_analysis}). We observe that perplexity margins drastically reduce as the context length increases, which offers insights into why acceptability judgement capability of the model improves/degrades with the choice of prefix.
Our results, therefore, can only be explained by the presence of implicit, instruction-free in-context learning ability of the model, and invite further scrutiny and investigation to long-form sentence understanding capabilities of LLMs.

\section{Background}
\label{sec:background}

\paragraph{Sequence Length and Out-of-domain Generalization.} When evaluating language models' linguistic abilities in particular, one ought to additionally consider the \emph{domain} of the test data fed into the model, as it can have large consequences for model performance if it mismatches from the model training data. Mismatching sequence lengths between (pre-)training and testing scenarios is well known to affect performance \citep{hupkes-etal-2020-compositionality, newman-etal-2020-eos, varis-bojar-2021-sequence}. As a simple example, the test pairs in standard \mpp{} datasets for the linguistic acceptability task are often fairly short (e.g. $\approx$ 4--30 tokens in the case of BLiMP). Because these test sequences are considerably shorter than that of the inputs LLMs typically receive during pre-training ($\approx$ 512--1024 tokens), we aim to investigate the extent to which LLMs' performance on acceptability judgements needs to be contextualized against work in length extrapolation, and in particular generalization during test time to both shorter and longer sequences.

\paragraph{Priming Language Models.} 
Recent work has explored the effects of providing additional linguistic context {to LLMs by ``priming'' or prepending their inputs with words/sentences.\footnote{This is related to but differs from the operationalization of priming as finetuning/adaptation as developed by \citet{van-schijndel-linzen-2018-neural, prasad-etal-2019-using}}
For instance, \citet{misra-etal-2020-exploring} and \citet{kassner-schutze-2020-negated} show LLMs to demonstrate semantic priming, assigning greater probabilities to words that were semantically related to words/sentence prefixes.
More recently, \citet{sinclair2022structural} used the priming paradigm to measure the probability assigned by LLMs to sentences when they are prefixed by well-formed but structurally different sentences.
They found LLMs to assign greater probability to sentences that are similar in structure to their prefixes across a number of diverse constructions, thereby demonstrating structural priming.}
Together with the findings of \citet{van-schijndel-linzen-2018-neural, prasad-etal-2019-using}, this suggests that LLMs can recognize and represent at least some of the relevant structural similarities between sentences.
While these methods do not focus on length \emph{per se}, their manipulation of the input context is necessarily accompanied by an increase in length. This leaves open the question as to how structural effects in context may interact with varying levels of input lengths, which we address in this work.
\paragraph{In-context Learning.}
A practical application of the priming paradigm is that it can be used to elicit learning behavior in LLMs. That is, LLMs can be primed using labelled task demonstrations \citep{brown2020language}, instructions/explanations \citep[though see \citeay{webson-pavlick-2022-prompt}]{lampinen2022can}, or a combination of the two \citep{wei2022chain,kojima2022reasoners} as supervision for tasks such as sentiment analysis or reasoning. This suggests that LLMs seem to be able to extract higher-level information from their context when processing a new test example of a supervised task. Our approach contributes to this body of work by testing if higher-level features for unsupervised tasks such as grammaticality can similarly be extracted by LLMs, given enough priming examples.


\section{Approach}
\label{sec:approach}


In this section, we describe the methods we use to probe the acceptability judgement perception of large language models with respect to change in the input length. 
\paragraph{Terminology.}
We follow standard practice in \mpp{}, where we evaluate the \textit{preference} ($\mathcal{P}$) of a language model $M$ towards acceptable sentence ($x$) over its unacceptable counterpart ($x')$, with respect to log likelihood, and compute the value over the full evaluation dataset $D$. $D$ typically consists of several \textit{test suites}, each of which instantiates a particular linguistic phenomenon. 
We denote the particular test suite under evaluation as the \textit{target suite}: $S \subset D$. Each target suite consists of $k$ pairs of acceptable and unacceptable sentences, $ (x,x')_{i=0}^k \in S$, and may have multiple conditions.
Within each target suite, we compute the acceptability judgements on one or more \textit{experimental conditions}, comparing a given LM's log-likelihood preference $\mathcal{P}$ for the acceptable and unacceptable sentence in each condition.
The accuracy score ($\mathcal{A}$) over a test pair from a single condition is defined as:
\begin{equation}
     \mathcal{A}(x_i, x_i') = \mathbbm{1}[\mathcal{P}(x_i) > \mathcal{P}(x_i')],
\end{equation}
\label{eq:baseline}
where $\mathbbm{1}$ is the indicator function which returns 1 if the inequality is satisfied and 0 otherwise.
Depending on the dataset, it can have either one or multiple conditions evaluated for each test item. 

To simulate increasing length of input, we prepend a prefix sequence $c_i$ to both $x$ and $x'$, and compute the preferences over the concatenated sequence, $\mathcal{P}([c_i, x_i])$ and $\mathcal{P}([c_i, x'_i])$, where $c$ can be of arbitrarily large length.



\paragraph{Datasets.} We use the standard targeted syntactic evaluation datasets of BLiMP \cite{warstadt-etal-2020-blimp} and SyntaxGym \cite{hu-etal-2020-systematic}. BLiMP is a large-scale \mpp{} dataset consisting of 67 different subsets of 1000 English sentence pairs each. Each BLiMP subset targets a separate linguistic paradigm that belongs to 12 different linguistic phenomena---for instance, \textit{subject-verb agreement}, \textit{argument structure}, etc. For each minimal pair of sentences $(x, x')_{i=0}^k$ in BLiMP, models are expected to rate the log-likelihood of the acceptable sentence $x$ above the log-likelihood of the unacceptable sentence $x'$.

SyntaxGym is a syntactic evaluation benchmark designed with more stringent evaluation criteria. For 34 different linguistic phenomena, the SyntaxGym benchmark defines test items with two to four different conditions, consisting of minimal structural variations on the same sentence which render the sentence either grammatical or ungrammatical. Model log-likelihoods are measured at a \emph{critical region} within each sentence, rather than across the whole sentence, and models are expected to produce log-likelihoods that satisfy multiple inequalities across all conditions. SyntaxGym is smaller than BLiMP (with about 20 items per phenomenon on average) and all of the examples are hand-written. We adapt 23 of the 34 test paradigms in SyntaxGym whose structure was compatible with the prefixing analyses of this paper. These two datasets offer complementary value to the analyses in this paper: BLiMP's large scale allows us to make general conclusions about the average effect of prefix interventions, while SyntaxGym's stringent evaluation allows us to verify that the effects are sustained under more rigorous experimental conditions.

\paragraph{Method.} 
We compute the log-likelihood on the given input using the \texttt{minicons} library \cite{minicons}, which is based on \texttt{huggingface} \cite{wolf-etal-2020-transformers}.
For each dataset $D$, we first compute the baseline acceptability accuracy according to \autoref{eq:baseline}.
Next, we aim to re-evaluate the acceptability accuracy by steadily increasing the token length of the input.
Following prior work on priming (\S\ref{sec:background}), we analyze how prepending the test examples with additional context affects a given model's acceptability judgements.
To increase the token length while maintaining the \mpp\ formulation, we introduce a context $c$ by prepending the same sequence to each target $x$ and $x'$ in $S$.
We also gradually increase the length of the context $c$ by sampling multiple sentences from a known set, and concatenating them using delimiters of periods and single spaces.

To construct a context $c$ we sample from several possible sources (acceptable sentences, unacceptable sentences, and control sentences), which we will discuss in more detail below. 
Then, we recompute the log-likelihood over the stimuli ($x$ or $x'$) by conditioning on $c$, i.e., $\mathcal{P}([c_i, x_i]) = \log p(x_i \mid c_i)$. 
For each item pair $(x_i, x'_i)$ in target suite $S \in D$, we first sample \textit{acceptable} sentences to construct $c$ from the following groups:

\begin{itemize}
    \item \textit{In-Domain}: Contexts are sampled from the same test suite as the target suite $S$: $x, c \in S, \ni x \ne c$.
    \item \textit{Out-of-Domain}: Contexts are sampled outside the target suite $S$: $x \in S, c \in D \ni c \notin S$.
\end{itemize}


For each $x \in S$, we construct $c$ by sampling $N$ sentences (without replacement) from each group described above, until the input reaches 1000 tokens\footnote{Since both GPT and OPT models support 1024 tokens in the context window, we opt to limit our investigation to 1000 tokens.}.

Traditionally, most work on priming has only considered acceptable sentences as the context. While there has been some work on syntactic priming in humans showing they can be primed with ungrammatical sentences to produce other ungrammatical sentences \citep{kaschak-glenberg-2004-construction, pickering-garrod-2017-priming, yang-etal-2019-syntactic}, there is little evidence in the NLP literature about how a model would react to \textit{unacceptable} sentences in the input.
Therefore, we perform our length evaluation on both acceptable prefixes ($c \in x$) and unacceptable prefixes ($c \in x')$, drawn from the same domain ($c \in S$) or from a different domain ($c \notin S$).

For evaluation, for each model, we compute the \textit{baselined accuracy} of acceptability judgements:
\begin{equation} \label{eq:3}
    \frac{1}{|D|} \sum_i^{|D|} A([c_i, x_i], [c_i, \hat{x_i}]) -  \frac{1}{|D|} \sum_i^{|D|} A(x_i, \hat{x_i}),
\end{equation}
where $|D|$ is the total number of samples in a given dataset ($D$). Taking this difference allows us to quantify the precise contribution (in terms of the gain or loss in accuracy of the LM on the acceptability task) of the priming contexts ($c$), which are held constant for a given pair of test samples. It further allows us to report a unified measure across our systematic manipulations of the context, as described above.

\paragraph{Models.} We investigate the length acceptability effect over autoregressive language models with varying scales---we consider GPT2 \citep{radford-etal-2018-improving}, and a subset of the OPT family (models including 125M, 350M, 1.3B, 2.7B and 6.7B parameters, \citealt{zhang2022opt}).

\paragraph{Control.}
While we define in-domain and out-of-domain with respect to the grammatical phenomena provided by the dataset (target suite, $S$), we are still in the regime of \textit{in-distribution} prefix sentences, as the context is drawn from the same \mpp\ dataset.
By design, these sentences are lexically constrained, and constructed to be as simple and as pared down as possible while still testing for the relevant phenomena.
To simulate \textit{out-of-distribution} context, i.e. contextual sentences having low similarity with the stimuli, we use Wikipedia data as a control case.
We sample prefixes from a completely unrelated and dissimilar domain, the WikiText-103 test set \citep{merity-etal-2016-wiki103}, to construct prefixes.
This test set is typically only used for evaluating large language models, so it should be out-of-distribution relative to both the training data of the language models, and to the \mpp{} sentences.

\paragraph{Regression Analysis.}

We define and test our claims about the effect of length on acceptability with a mixed-effects logistic regression for each combination of model and dataset. The regression predicts a model's acceptability judgement accuracy for a given task suite as a function of the three properties of the prefix $c$ we introduced previously: its length, whether it is in-domain or out-of-domain, and its acceptability. The model includes a three-way interaction term and all lower-order terms for these variables, with sum-coded categorical variables and log-transformed prefix lengths, along with a random intercept term for the task suite (controlling for variation in baseline accuracies per suite).

\section{Main Results}\label{sec:results}

\begin{figure*}[t]
    \centering
    \includegraphics[width=0.8\linewidth]{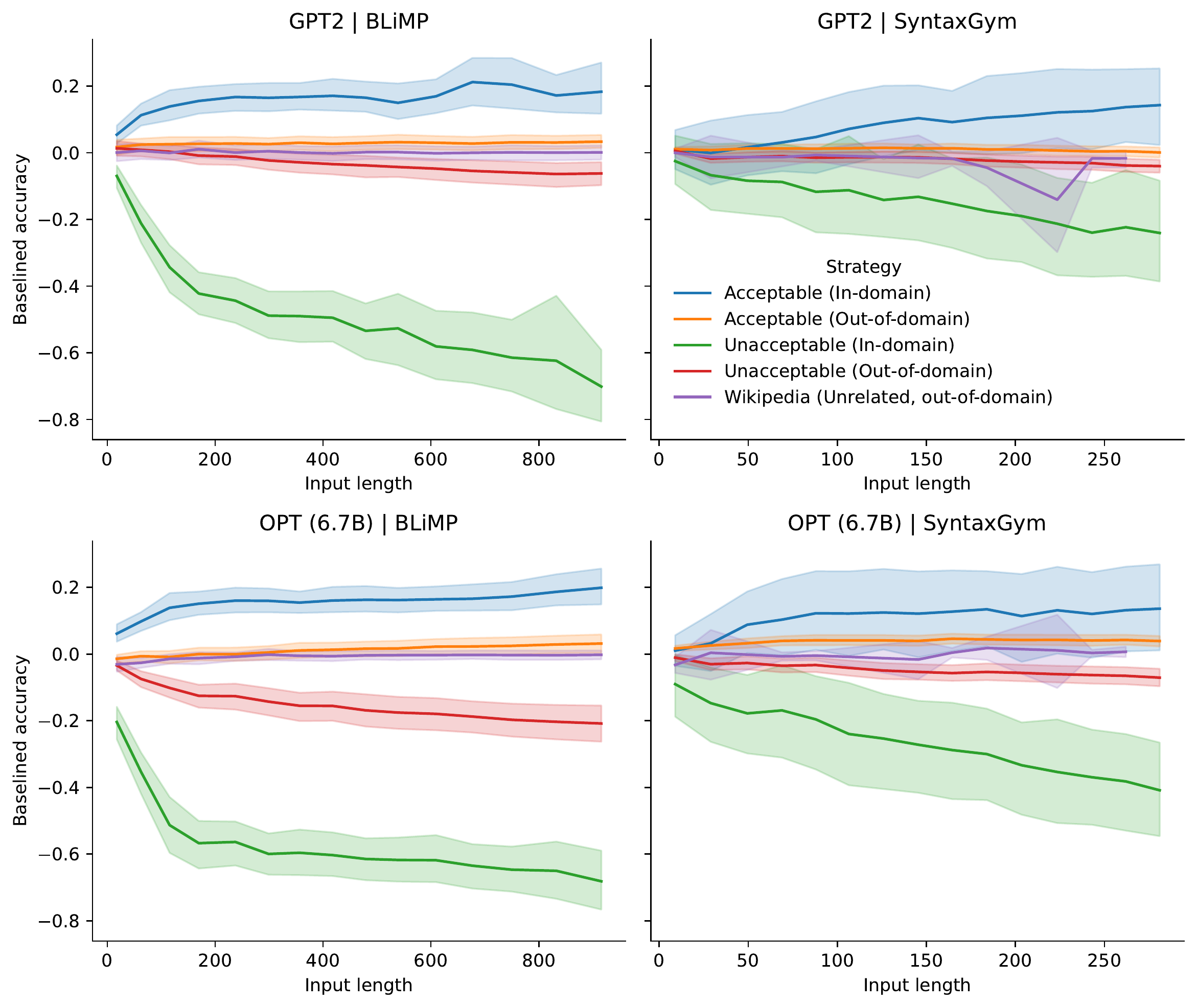}
    \caption{Average effect of prefixing targeted syntactic evaluations with acceptable or unacceptable content drawn from the same syntactic evaluation (in-domain) or a different syntactic evaluation (out-of-domain).}
    \label{fig:all-combined}
\end{figure*}

\newcommand\pvalLengthPrefixType{($p<0.002$ for all models on BLiMP and SyntaxGym)}
\newcommand\pvalBlimpLengthPrefixType{($p < 10^{-11}$ for all models)}
\newcommand\pvalLengthPrefixTypeDomain{($p < 0.007$ for all models on BLiMP and SyntaxGym)}
\newcommand\pvalSGLengthPrefixTypeDomain{($p < 0.007$ for all models)}
\newcommand\pvalBlimpLengthPrefixTypeDomain{($p < 10^{-4}$ for all models)}
\newcommand\pvalWikiLength{($p > 0.2$ for all models)}
\newcommand\pvalBlimpWikiLength{($p > 0.4$ for all models)}
\newcommand\pvalSyntaxGymWikiLength{($p > 0.2$ for all models)}

\begin{figure}[t]
    \centering
    \resizebox{\linewidth}{!}{
    \includegraphics[]{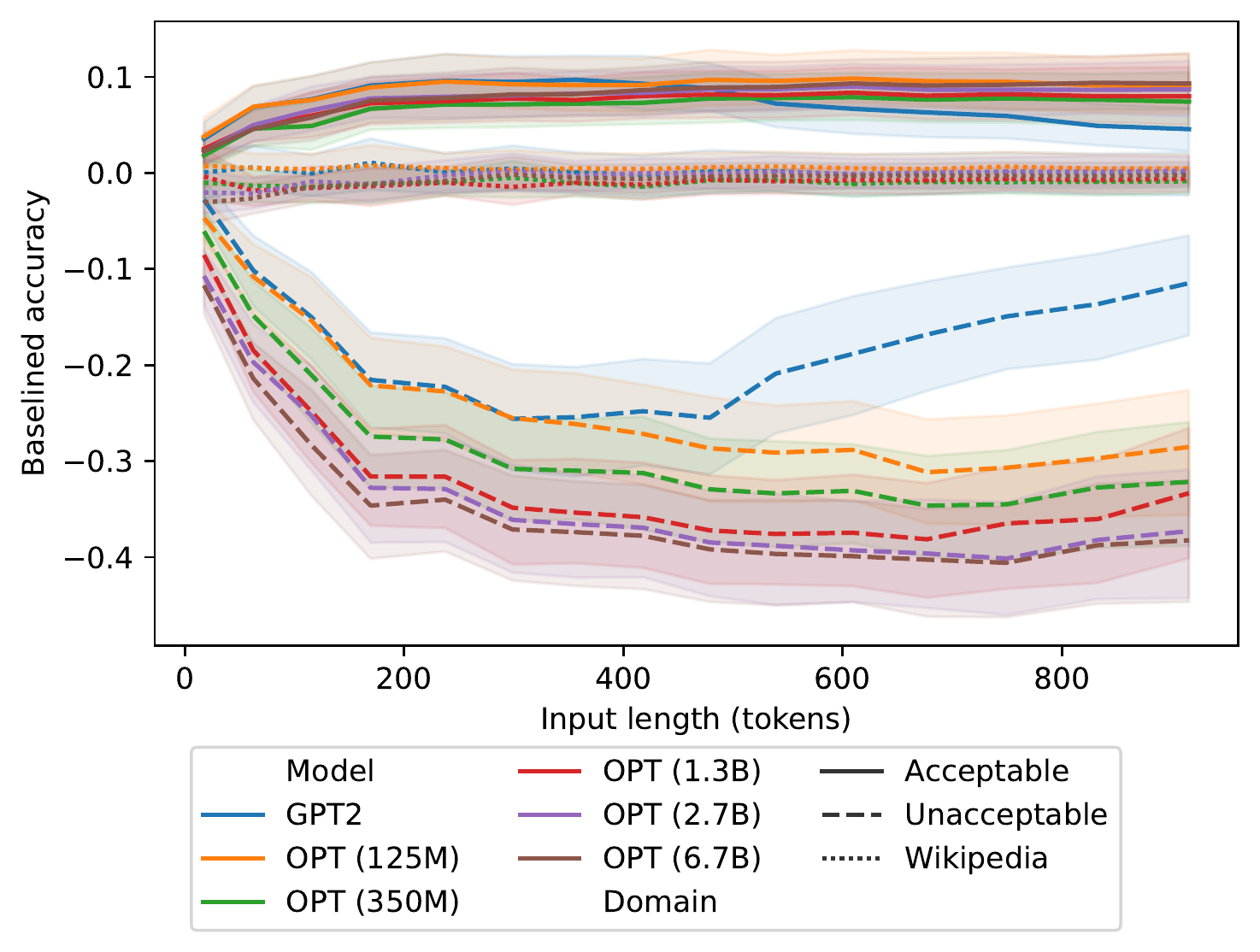}}
    \caption{Interaction of length and prefix type on BLiMP acceptability judgements. Across all tested models, accuracy improves for acceptable prefixes and worsens for unacceptable ones, as length increases \pvalBlimpLengthPrefixType. Shaded regions describe the 95\% confidence interval.}
    \label{fig:blimp_positive_effect}
\end{figure}

\autoref{fig:all-combined} presents the summary results of our prefixing manipulation, charting models' baseline accuracy on \mpp{} evaluations as a function of properties of the prefixed content: (1) its length (x-axis), (2) its acceptability (blue and orange vs. red and green), and (3) whether it is drawn from an unrelated domain, Wikipedia (purple), same domain (blue and green) or a different domain (orange and red). We walk through the main qualitative findings in the following paragraphs.






\paragraph{\textit{The length of the input impacts the acceptability judgement, depending on the nature of the prefix.}} 


We first observe the impact of increasing context length on the acceptability judgements of the models. As we prefix longer grammatical content, models monotonically improve in average accuracy (\autoref{fig:blimp_positive_effect}, dashed lines). Model accuracy increases up to 20 percentage points, and mostly uniformly across all model sizes. Simultaneously, we observe a strong negative effect of using ungrammatical prefixes: acceptability reduces sharply for models with an increase in context length. Quantitatively, this interaction between prefix length and acceptability of prefix is highly significant in all models and evaluations \pvalLengthPrefixType.

This negative effect is amplified in larger models. For example, OPT 6.7B suffers the largest degradation of acceptability with increasing length of ungrammatical context (\autoref{fig:blimp_positive_effect}, solid lines).%
\paragraph{\textit{In-domain context impacts acceptability judgements more than out-of-domain contexts.}}

\begin{figure}[t]
    \centering
    \resizebox{\linewidth}{!}{
    \includegraphics[]{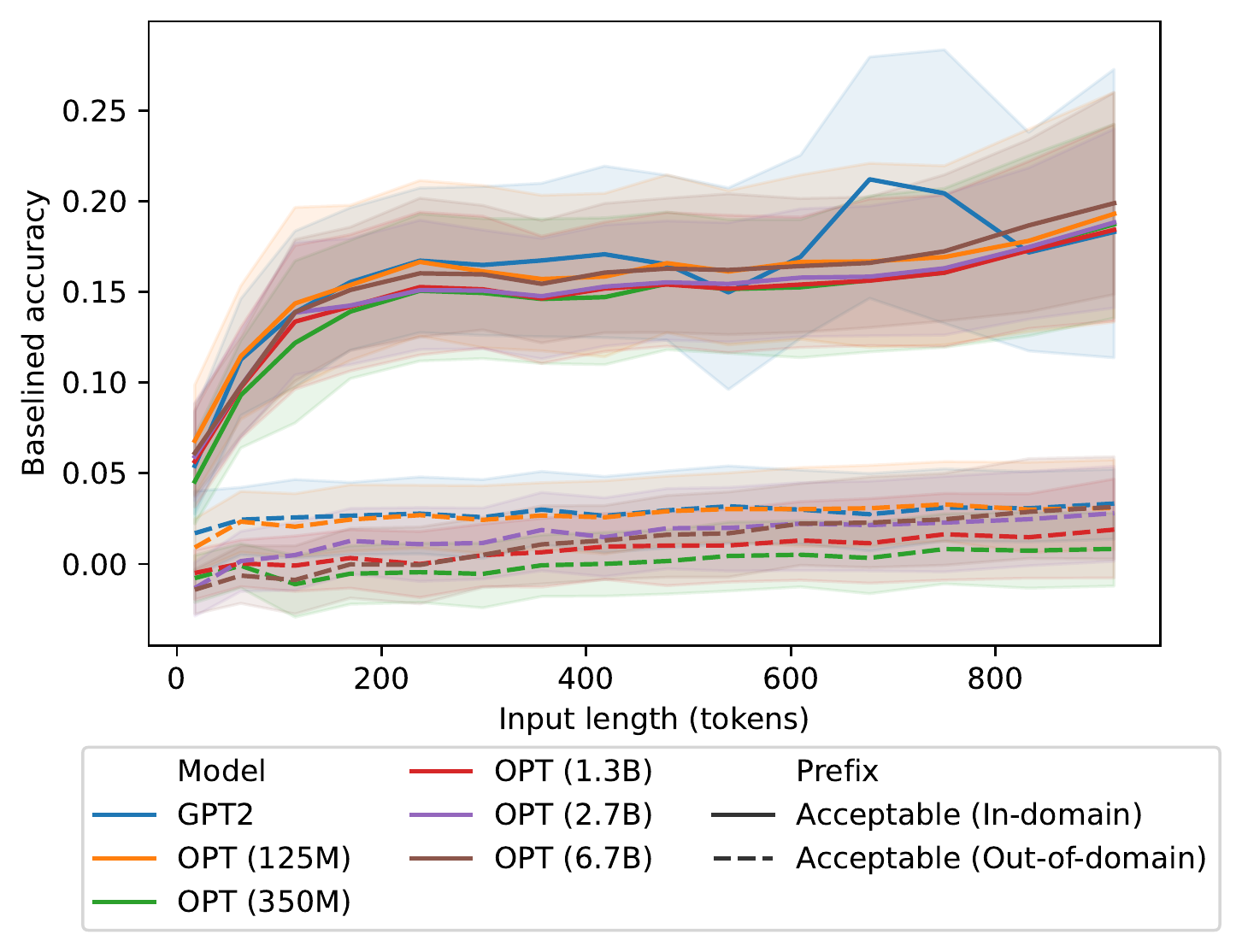}}
    \caption{In-domain vs Out-of-domain grammatical context input effect of length on acceptability judgements.}
    \label{fig:blimp_invsout_grammatical}
\end{figure}


\begin{figure}[t]
    \centering
    \resizebox{\linewidth}{!}{
    \includegraphics[width=\linewidth]{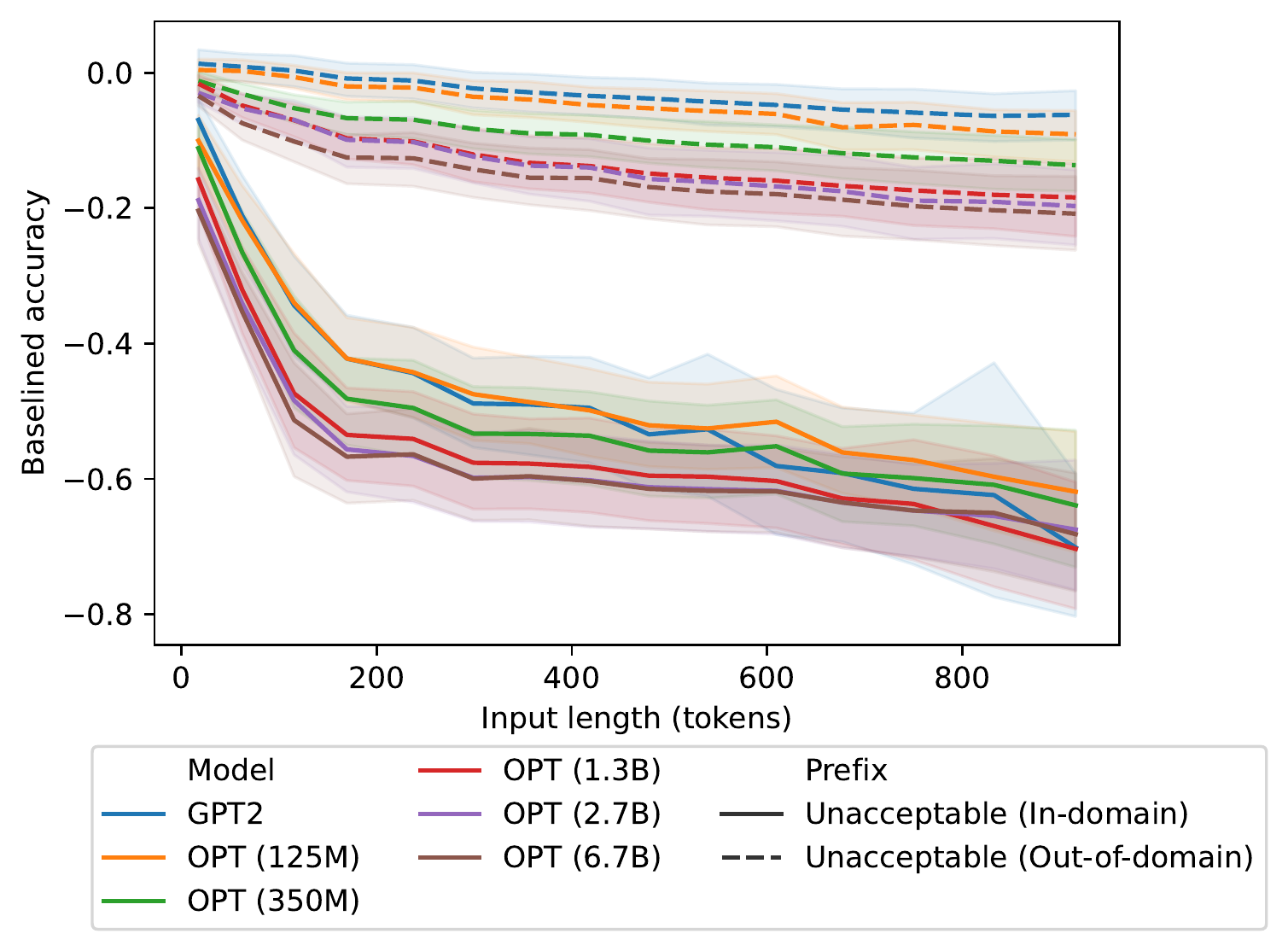}}
    \caption{In-domain vs Out-of-domain ungrammatical context input effect of length on acceptability judgements.}
    \label{fig:blimp_invsout_ungrammatical}
\end{figure}

We now investigate the previous result closely with respect to the similarity of the context to the stimuli while increasing length. In other words, we tease out the effect of \textit{in-domain} and \textit{out-of-domain} contexts with increasing context length. 
We observe sharp improvement in acceptability judgements in BLiMP (\autoref{fig:blimp_invsout_grammatical}) for in-domain compared to the out-of-domain contexts, with a baselined accuracy increase of up to 20 percentage points. 
Out-of-domain contexts account for only a marginal improvement in acceptability.
However, we note that the improvement does not correspond to the increase in model size (in terms of number of parameters).


Next, we observe even larger impact of in-domain contexts when priming with ungrammatical contexts. 
The acceptability score plummets drastically with increasing input length. With an increase in the scale of model parameters, we observe more amount of drop with respect to baselined accuracy (\autoref{fig:blimp_invsout_ungrammatical}), with a reduction of more than 70\% percentage points for OPT 6.7B at token lengths greater than 700.
Similarly, we also observe a drop in performance in judging acceptability when unacceptable prefixes are used from out-of-domain, but the amount of performance drop is relatively less compared to in-domain scenario. OPT 6.7B parameter model displays the biggest drop, but up to 20 percentage points, compared to 70 percentage points for in-domain case. 
These results suggest that on average, LMs tend to prefer unacceptable continuations when prefixed with sentences that were also unacceptable. This behavior is importantly amplified by a substantial amount when the unacceptability of the prefixes was consistent---i.e., when they violate the grammatical rules (of English) in the same way (in-domain), as opposed to in more diverse ways (out-of-domain).
This is not necessarily a negative result, as it lends support to the conjecture that LMs are sensitive to the abstract ``acceptability'' feature that is emergent in their input. 

Overall, this experiment shows that the effect of priming monotonically increases with an increase in the context length, and the domain of the phenomena has a large impact on the performance in judging acceptability of sentences.
These effects manifest quantitatively in a three-way interaction between prefix acceptability, in-domain vs. out-of-domain, and prefix length \pvalLengthPrefixTypeDomain.
\paragraph{\textit{Negligible effect on acceptability with an increasing length of unrelated prefixes.}}


Our control experiments using prefixes drawn from Wikipedia do not display a significant change in the acceptability of the models under investigation, suggesting that irrelevant context has hardly any effect on priming (see purple lines in \autoref{fig:all-combined}).\footnote{Note, however, that we have assumed that (i) Wikipedia sentences will be acceptable, and (ii) that acceptable prefixes had a generally weaker effect on the acceptability task. If we were to test \emph{un}acceptable Wikipedia sentences as well, we could have seen a small priming effect. How to best generate ungrammatical data in the Wikipedia domain isn't a trivial exercise, so we leave this avenue of investigation for future work.}
In a separate regression analysis, we predicted the performance of models' acceptability judgements on BLiMP for either grammatical BLiMP prefixes (both in-domain and out-of-domain) or prefixes drawn from Wikipedia. The regression predictors included the prefix length, whether the prefix is drawn from BLiMP or Wikipedia, and their interaction. No effect of prefix length is significant \pvalWikiLength{} for Wikipedia sentences.
This result also reinforces the findings from \citet{sinclair2022structural}, as Wikipedia sentences doubtless have the least structural overlap with the BLiMP and SyntaxGym task suites.

\section{Prefix Similarity Analysis}
\label{sec:similarity_analysis}

We have observed that length effects on acceptability judgements are conditional on the similarity of the prefix to the test sentence. Does some specific kind of similarity (e.g., syntactic or lexical similarity) explain this phenomenon? Perhaps the prefix is syntactically priming the model for the target sentence \citep{sinclair2022structural}, in which case we would expect the syntactic similarity of the sentences to correlate with accuracy when using grammatical prefixes. Another possibility is that a more spurious feature---such as lexical overlap---is responsible \citep{misra-etal-2020-exploring, kassner-schutze-2020-negated}. To test this, we can correlate syntactic similarity and lexical overlap with accuracies on each example.

To measure lexical overlap, we use $F_1$ scores to measure how many tokens\footnote{We tokenize the inputs using GPT2's tokenizer before computing overlap.} in the prefix and test sentences are shared. To approximate syntactic overlap, we can compute the $F_1$ score over \textit{dependency labels} in two sentences, rather than across tokens. If multiple prefix sentences are present, we can take the mean similarity with the target sentence across prefixes. Then, we compute the point-biserial correlation\footnote{The point-biserial correlation coefficient measures the strength of the relationship between a continuous variable (e.g., our overlap metrics) and a binary variable (accuracy on an individual example).} ($\rho_p$) between the similarity metric and accuracy on a given example, averaging similarities across prefix sentences. We compute the correlation separately for each model size and each prefixing strategy. Note that we only use grammatical prefixes; thus, we expect positive correlations if priming explains the length effects. We find very low and non-significant correlations with dependency overlap and token overlap ($\rho_p < 0.05$, $p > 0.1$) regardless of prefixing strategy or model size. This could be evidence that the model is more sensitive to the length of the prefixes than any notion of syntactic or lexical similarity on this task.

However, this instance-level analysis could be confounded by the mixture of various phenomena in the prefixes. The model could be sensitive to sentences from certain phenomena more than others, or the varying lengths of sentences from each phenomenon. To more specifically measure whether priming can explain our findings, we focused on BLiMP and prefixed sentences from one phenomenon at a time with a given test phenomenon; in other words, we sample \textit{out-of-domain} prefixes, but controlling which phenomenon we sample from. Using this approach, we can capture how structurally similar each BLiMP phenomenon is with each other BLiMP phenomenon, and how this correlates with accuracies. We describe the out-of-domain single-phenomenon prefixing strategy and present similarity metrics in Appendix~\ref{app:similarities}. When correlating phenomenon-level similarities with accuracies, we find that correlations here are a bit stronger than when we mix out-of-domain prefixes ($\rho_s = 0.11$ for dependency overlap, and $\rho_s = 0.18$ for token overlap, $p < 0.001$ for both). The magnitude of the correlations is low, but they are still significant. Thus, there is some relationship between the similarity of the prefix and test sentence with accuracy, but the relationship is weak.

This is preliminary evidence that lexical overlap and low-level syntactic similarity effects \emph{partially} explain accuracy increases with BLiMP prefixing, but most of the trends we observe cannot be explained by these effects alone. Perhaps this is because the model is more sensitive to multiple similarities simultaneously than any one isolated type of similarity. Or, perhaps models are sensitive to some other latent feature that we did not analyze. Nonetheless, it is difficult to draw strong conclusions from the lack of a strong correlation, and correlations alone cannot causally implicate similarities in explaining our findings. 
Perhaps future work could disambiguate the relationship between these factors using causal methods.










\section{Discussion}

Our analyses have revealed that, on average, language models' acceptability judgements are highly sensitive to the domain and acceptability of the input in their contexts. This has implications for interpreting results from \mpp{} benchmark datasets: single-sentence or otherwise short inputs may not be representative of models' true abilities. Indeed, shorter inputs may not be what pre-trained language models expect, given that their pre-training procedures often entail packing many sentences into a single training example. This agrees with prior work that finds performance improvements from reformatting train and test inputs in a way that more closely resembles the pre-training setup \citep{hupkes-etal-2020-compositionality,newman-etal-2020-eos,varis-bojar-2021-sequence,chada2021fewshotqa}. 

Crucially, however, \textbf{performance is only sensitive to length given \emph{in-domain examples}}.
Our results also demonstrate a notable capacity of LMs to show behavior that is consistent with the acceptability of their prefix. That is, while models showed marked improvements on acceptability tasks when prefixed by acceptable sentences from the same domain, they also (more substantially) showed the opposite behavior---preferring unacceptable sentences---when prefixed by sentences that were unacceptable \textit{in the same way}. This two-way consistency adds more credence to the general observation that models tend to be sensitive to abstract features (such as acceptability) emergent in their context, than recent work that only explores this behavior in one direction \citep{lampinen2022nested, sinclair2022structural}.

More broadly, this adds to the literature on prompt sensitivity in pre-trained language models. Prompt tuning work has found that LMs are sensitive to individual prompts \citep{kojima2022reasoners}, and that the ordering of in-context examples \citep{lu2022fantastically} greatly affects model performance. Smaller LMs are also sensitive to the choice of prompt and output verbalizer \citep{schick2021exploiting,gao2021making}, and we indeed find sensitivity to prefixes in our study for a variety of model sizes and prefixing strategies. To our knowledge, however, our study is the first to implicate input \emph{length} as a factor contributing to linguistic performance. Practically, the length effects we find may be more significant in the presence of in-distribution training examples or in-domain prompts when using in-context learning/prompting. Future work could verify length effects across various downstream tasks.

Our model-based similarity analysis in \S\ref{sec:similarity_analysis} and Appendix~\ref{app:similarities} did not demonstrate a clear causal story for length effects on performance; indeed, a bottom-up study of model judgements for individual syntactic phenomena would be required to better understand why this effect holds. We did find that the lexical similarity of the prefixes with the test input correlated more strongly than syntactic similarity with accuracies, though the strength of the correlations was small (but still significant). Appendix~\ref{app:syntaxgym-cross-priming} contains details on individual suites' performance in our prefixing methodology, and a brief discussion of salient trends in the data. Thus, lexical and low-level syntactic similarity effects cannot be directly implicated in length effects on performance given our analyses, but in-domain examples \emph{do} nonetheless have a much stronger effect than out-of-domain examples. Future work could further investigate why models' acceptability behavior is more susceptible to contextual influence for some syntactic phenomena over others.

\section{Conclusion}

In this work, we perform a systematic study to the robustness of Transformer language models' syntactic acceptability judgements to manipulations of the judgement context. Specifically, we closely study how the grammatical preferences of a language model changes in the minimal-pair paradigm (\mpp), when the input to the model is arbitrarily long. 
To simulate the \mpp\ setup while increasing the input length, we propose a mechanism to introduce long contextual sentences to existing \mpp\ datasets, such as BLiMP and SyntaxGym, which consists of a number of grammatical phenomena for evaluation.

We find that model acceptability judgements are generally robust when placed in randomly sampled linguistic contexts, but that particular manipulations of the context can drive up or down the accuracy of their judgements. In particular, contexts containing syntactic structures which closely match those in the test sentence can improve or degrade the models' judgement performance, if those context sentences are acceptable (grammatical) or unacceptable (ungrammatical), respectively. This effect is amplified as we increase the length of the context provided to the model.
Our results demonstrate in-context learning in a highly specific way: models are sensitive to granular syntactic properties of the context when making predictions over a target sentence, such that they can be driven to produce both correct and reliably \emph{incorrect} outputs.


\section*{Acknowledgements} 
We would like to thank Marten van Schijndel, Allyson Ettinger, Tiwalayo Eisape, Jennifer Hu, Peng Qian and Alex Warstadt for their feedback and comments on this draft. 

\bibliography{arxiv,custom}
\bibliographystyle{acl_natbib}

\appendix

\section{BLiMP Phenomenon Similarities}\label{app:similarities}
\begin{figure*}
    \includegraphics[width=0.48\linewidth]{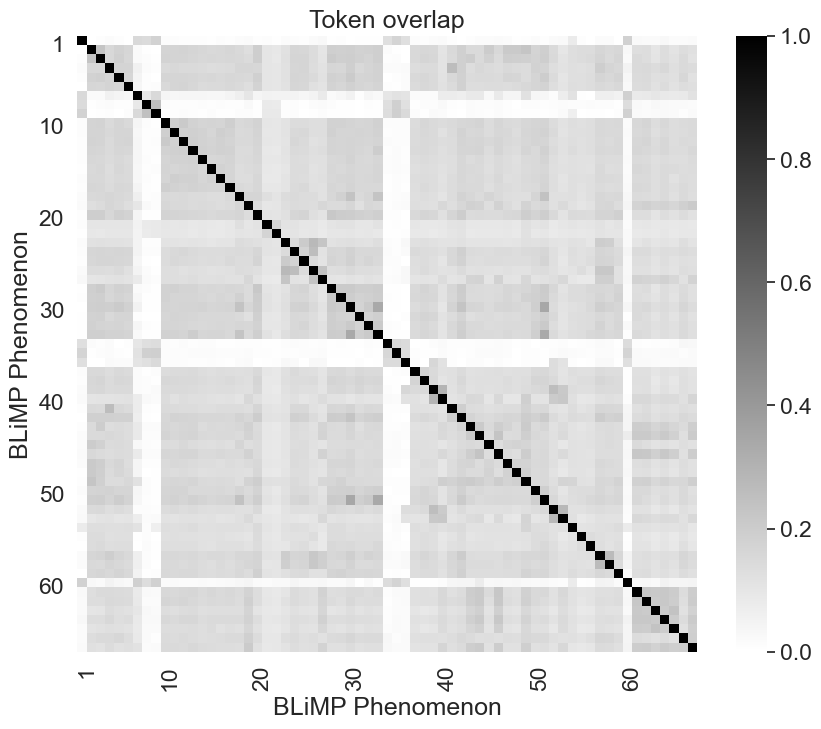}
    \hfill
    \includegraphics[width=0.48\linewidth]{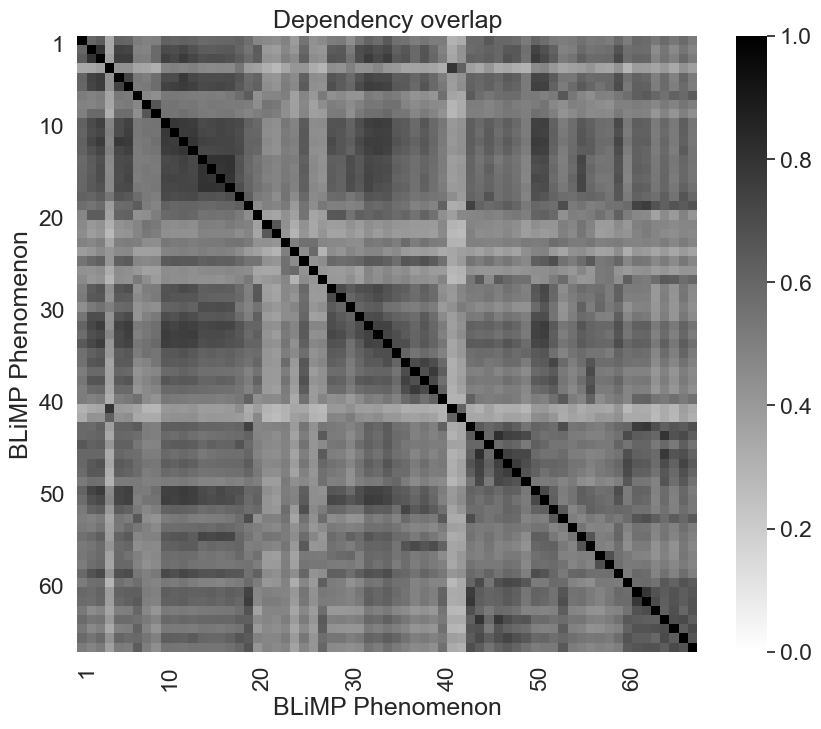}
    \caption{Token overlap (left) and dependency overlap (right) across BLiMP phenomena. We compute these using a sample of 10,000 sentences from the target phenomenon and from the prefix phenomenon. The phenomena are ordered alphabetically.}
    \label{fig:blimp_similarities}
\end{figure*}

\begin{figure*}
    \includegraphics[width=0.48\linewidth]{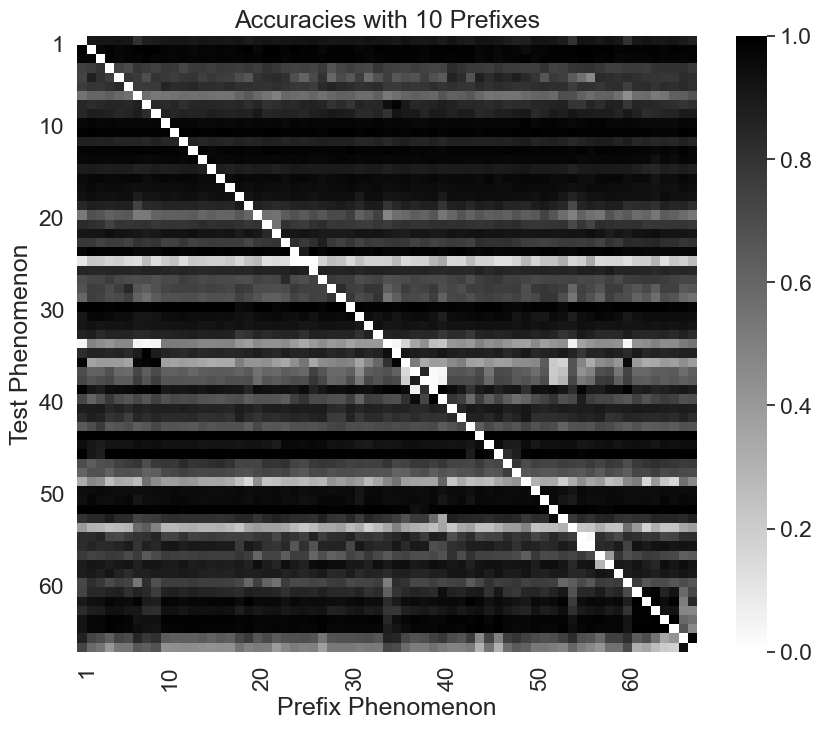}
    \hfill
    \includegraphics[width=0.48\linewidth]{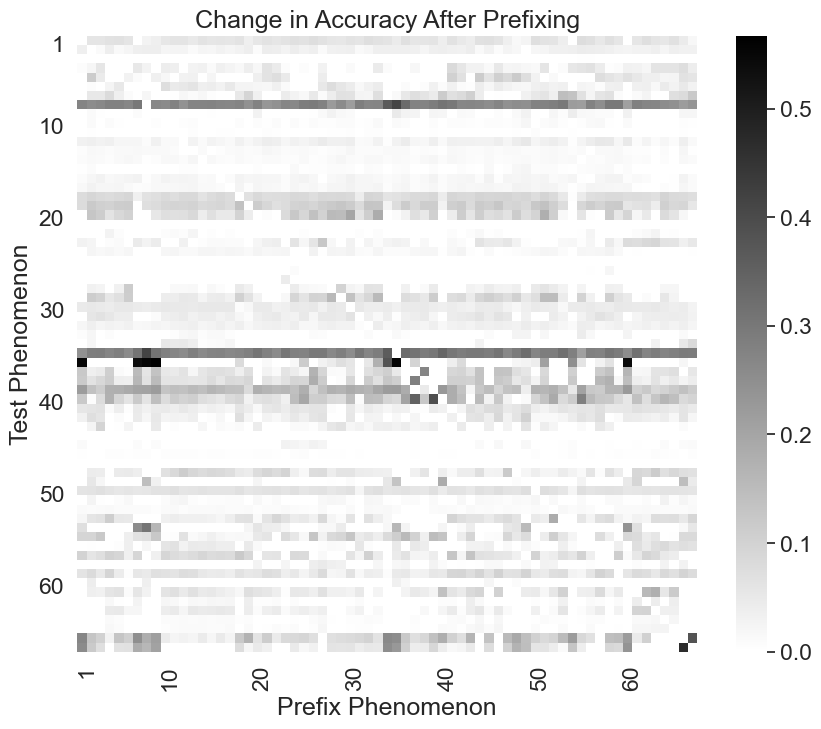}
    \caption{Accuracies for GPT2 on individual BLiMP phenomena after prefixing 10 sentences from a single BLiMP phenomenon (left). Change in accuracy from no prefix to 10 prefixes on each BLiMP phenomenon (right). We exclude the diagonal in both cases, as we are interested in \textit{out-of-domain} prefixing effects.}
    \label{fig:blimp_accs_by_phenomenon}
\end{figure*}

Here, we present the lexical and syntactic similarity across each pair of BLiMP phenomena (Figure~\ref{fig:blimp_similarities}).\footnote{For visual conciseness across confusion matrices, we use indices rather than individual phenomenon names. For each confusion matrix in Figures~\ref{fig:blimp_similarities} and \ref{fig:blimp_accs_by_phenomenon}, all phenomena are presented in alphabetical order.} These are computed across each prefix and test phenomenon using a sample of 10,000 test sentences and 10,000 prefix sentences for each point in the confusion matrix. We find that dependency overlap is generally higher than token overlap across inputs, perhaps unsurprisingly given that the size of the set of possible dependency labels is much smaller than the size of the set of possible tokens in a given sentence.

We next try correlating these values with accuracies on each BLiMP phenomenon as a function of these phenomenon-level similarity metrics. Accuracies with prefixes (and changes in accuracies after after prefixing) for GPT2 are presented in Figure~\ref{fig:blimp_accs_by_phenomenon}. Essentially, we are now measuring how similar the trends are across a similarity confusion matrix and an accuracy confusion matrix. As we are now measuring similarity across continuous variables, we compute the Spearman correlation ($\rho_s$). We find that correlations here are a bit stronger than when we mix out-of-domain prefixes ($\rho_s = 0.11$ for dependency overlap, and $\rho_s = 0.18$ for token overlap, $p < 0.001$ for both). While the magnitude of the correlations is very low, these are still significant. Thus, there is some relationship between the similarity of the prefix and test sentence with accuracy, but the relationship tends to be weak. Also, lexical overlap seems to be more strongly predictive of accuracies than structural similarities, indicating that the model may indeed be more sensitive to spurious lexical similarities than any deeper abstract notion of syntactic similarity between a prefix and the test sentence. Nonetheless, this is still preliminary evidence that priming effects do not explain much of the accuracy trends we observe with prefixing; instead, perhaps length itself makes a stronger difference than any specific notion of similarity between the prefix and test sentence.

While we find weak correlations between similarities and accuracies, perhaps lexical and structural similarities simply do not correlate with accuracies directly. Instead, the strength of the prefix's effect on accuracy with increasing length could correlate with similarity. In-domain prefixes have very high token overlap and 100\% dependency overlap with the test sentence, and these prefixes have a much larger positive effect on accuracy with increasing numbers of prefixes (when grammatical). Thus, perhaps future work could consider directly implicating priming effects in the strength of the accuracy change with more prefixes; this would require more than correlation coefficient metrics, but could provide interesting insights into the inner workings of how models leverage long prefix contexts when making predictions on a given test example.

\section{Suite-by-suite prefixing performance}
\label{app:syntaxgym-cross-priming}

\begin{figure*}
\centering
\includegraphics[width=\linewidth]{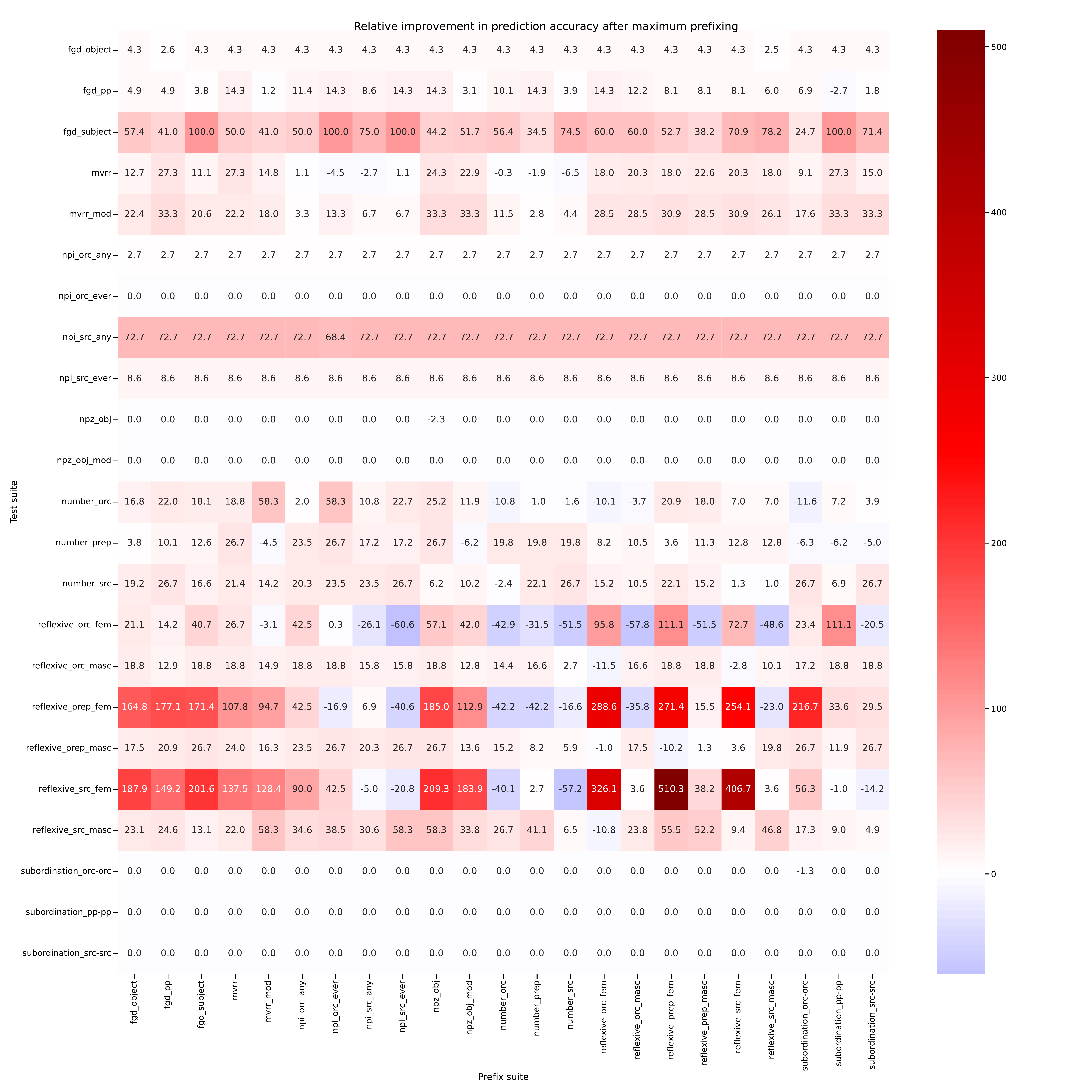}
\caption{Relative improvement (in percentage points) in accuracy on SyntaxGym test suite evaluations (rows) after prefixing with sentences from other SyntaxGym test suites (columns) for GPT2.}
\label{fig:syntaxgym-cross-priming}
\end{figure*}
\autoref{fig:syntaxgym-cross-priming} shows GPT2's improvement in prediction accuracy on different SyntaxGym test suites (rows) after drawing as many acceptable prefix sentences as possible from another SyntaxGym test suite (columns). The values are a percentage increase in prediction accuracy, relative to GPT2's baseline performance with no additional context. We see a substantial diversity in how different suites respond to prefixing of acceptable sentences. Some suites, such as an NPI licensing suite (\texttt{npi\_src\_any}) and a filler-gap dependency suite (\texttt{fgd\_subject}), show across-the-board improvements in response to any prefixing at all.
The suites labeled \texttt{reflexive\_*\_fem}, which test understanding of feminine reflexive anaphor agreement, demonstrate interesting unstable behavior: GPT2's predictions degrade when these particular tests are preceded by grammatical sentences containing masculine reflexive anaphors (see e.g. the blue boxes in the row labeled \texttt{reflexive\_orc\_fem}, but the same predictions are facilitated when preceded by feminine reflexive anaphors.

\section{Margin Analysis}
\label{sec:margin_analysis}

\begin{figure}
    \centering
    \includegraphics[width=\linewidth]{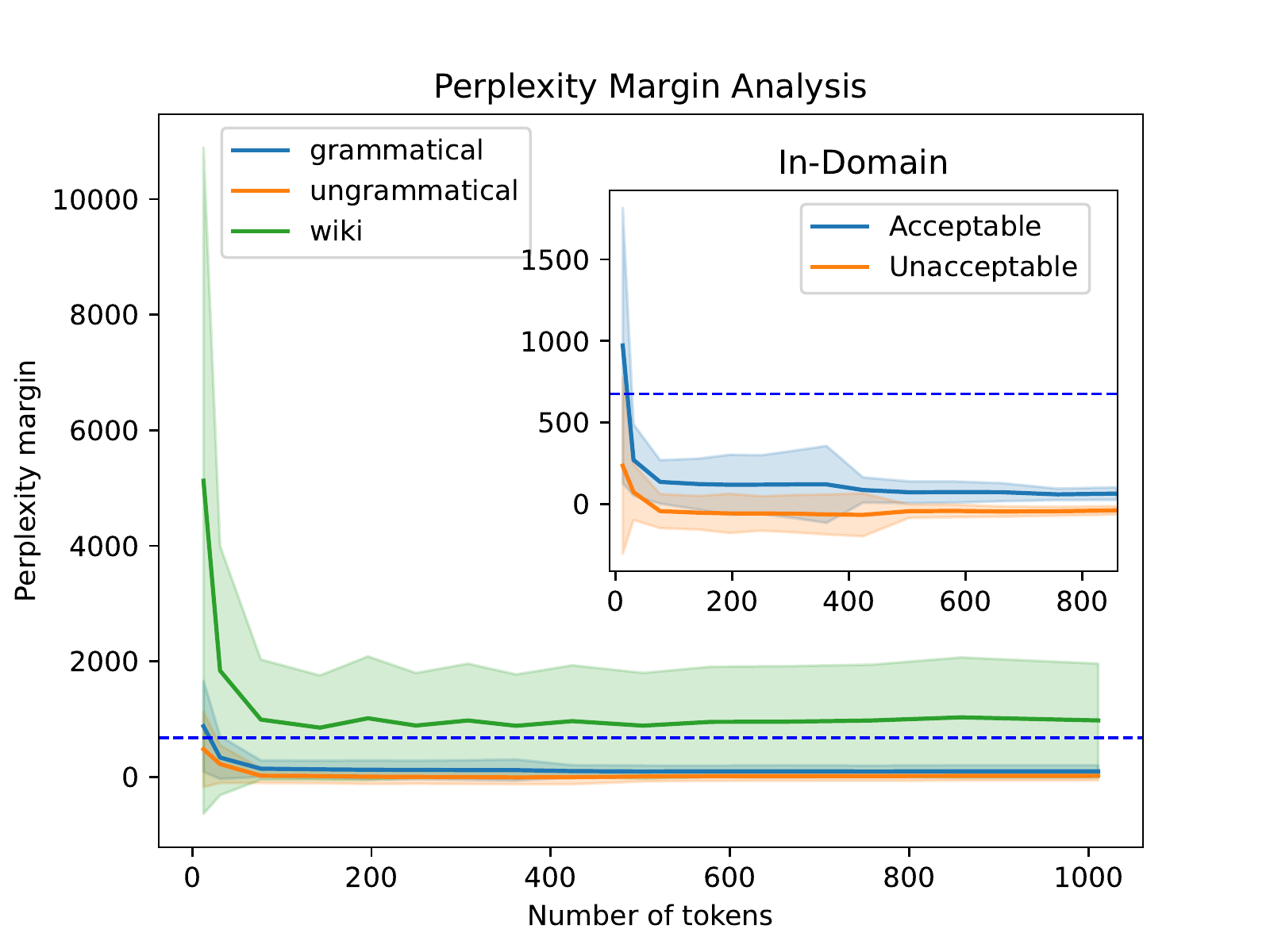}
    \caption{Perplexity margins of Grammatical, Ungrammatical and Wikipedia prefixes on BLiMP for OPT 6.7B model. The dashed lines represent the mean margin of the baseline without any context.}
    \label{fig:blimp_margin_wiki}
\end{figure}

How confident are LMs as input length increases? The results on length priming indicates that longer in-domain, acceptable prefixes tend to induce better acceptability judgements to the target model. 
However, investigating the accuracies as computed in \autoref{eq:3} alone does not fully explain the nuances of the confidence of the model. 
To understand how the confidences themselves differ in acceptable/unacceptable target sentences, we plot and investigate the perplexity margins in \autoref{fig:blimp_margin_wiki}. 
Specifically, we compute the difference in the model perplexities $\delta$ for each acceptable/unacceptable pair:

\begin{equation}
     \delta(x_i, \hat{x_i}) = \mathcal{P}(x_i) - \mathcal{P}(\hat{x_i}),
\end{equation}

We observe the margins on BLiMP for a candidate model, OPT 6.7B in \autoref{fig:blimp_margin_wiki}, for grammatical, ungrammatical and Wikipedia prefixes. For all cases, $\delta$ starts from a high value for short sequences, and approaches zero as the context length increases.
There is a marked difference in $\delta$ values compared to Wikipedia and BLiMP prefixes: Wikipedia prefixes appears to display a high value, suggesting high surprisals. Average $\delta$ for Wikipedia also remains higher than the baseline value (without any priming), while $\delta$ is significantly lower for BLiMP prefixes. 
This behavior potentially explains why we observe almost no change in the accuracy of Wikipedia prefixes, as the margin remains high and stable with increasing length of tokens.

Within the in-domain prefixes, we observe the $\delta$ to be significantly lower for unacceptable prefixes compared to the acceptable contexts, and it reduces with length. This behavior partially explains why we observe the trend of sharp decrease in acceptability accuracy for in-domain unacceptable prefixes, as the monotonically decreasing $\delta$ flips the acceptability judgement associations.



\section{SyntaxGym figures}

\begin{figure}
\includegraphics[width=\linewidth]{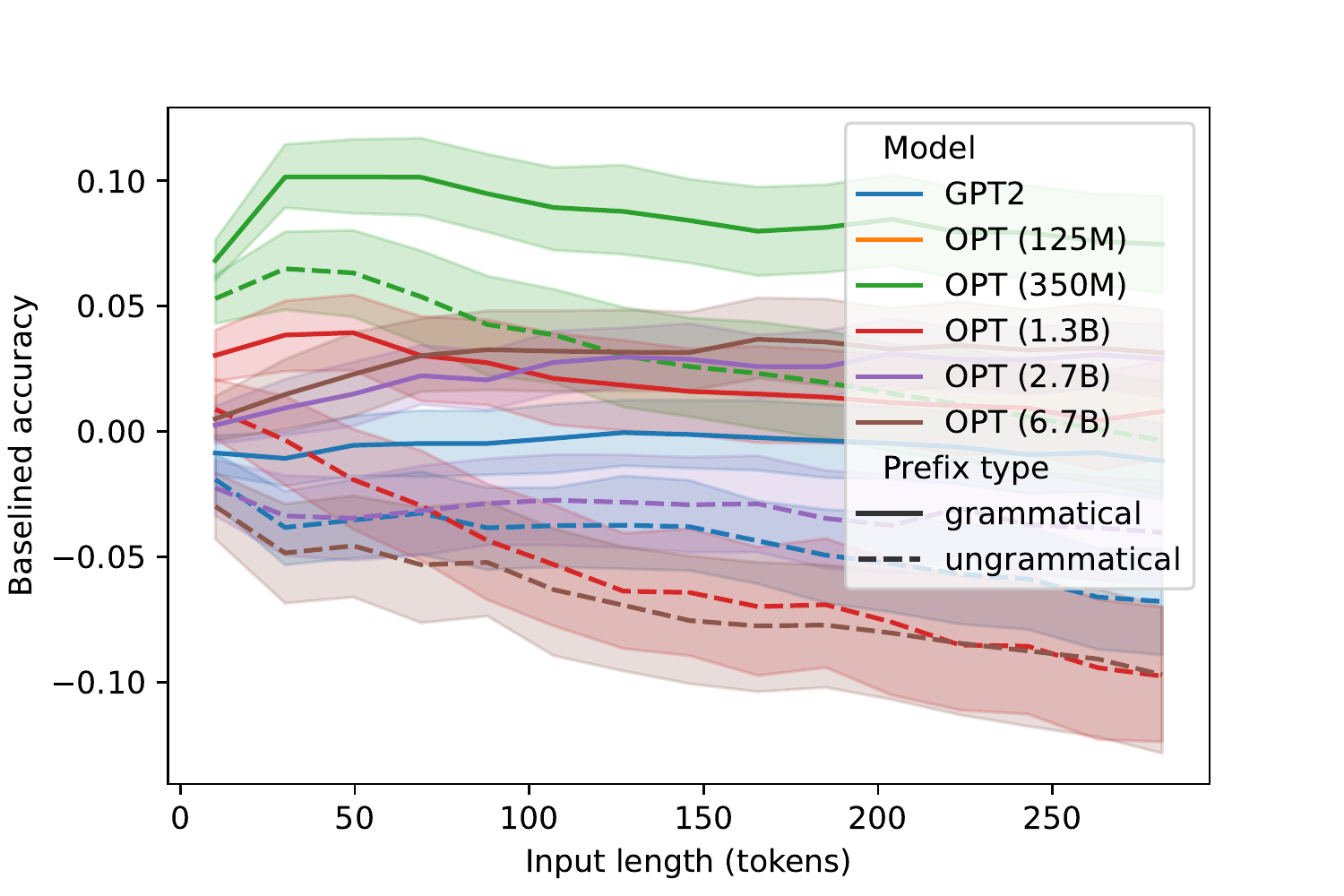}
\caption{Effects of length and grammaticality per model. Analogous to \autoref{fig:blimp_positive_effect} in the main text.}
\label{fig:syntaxgym-positive-grammatical}
\end{figure}

\begin{figure}
\includegraphics[width=\linewidth]{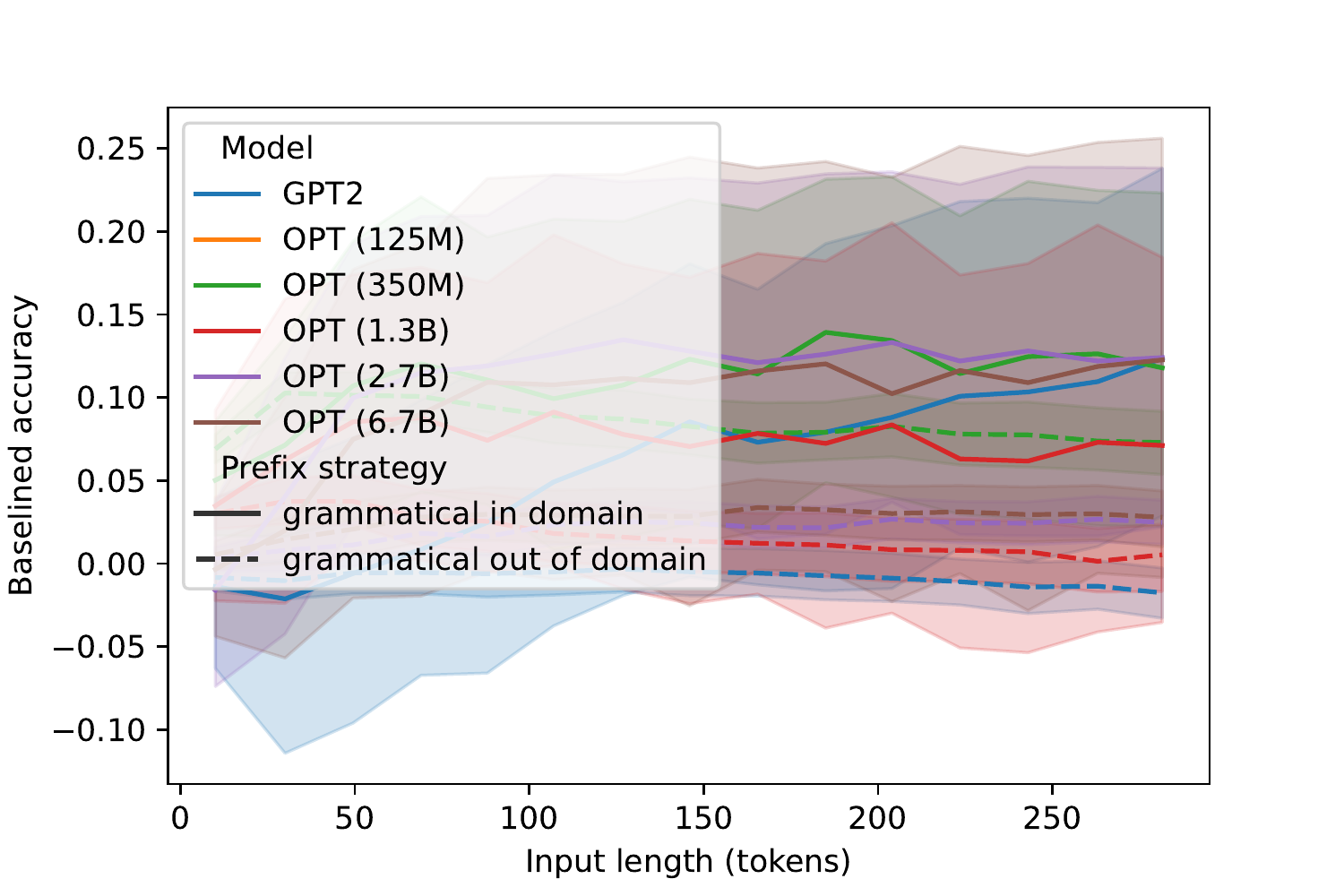}
\caption{Effects of length and domain per model for grammatical prefixes. Analogous to \autoref{fig:blimp_invsout_grammatical} in the main text.}
\label{fig:syntaxgym_invsout_grammatical}
\end{figure}

\begin{figure}
\includegraphics[width=\linewidth]{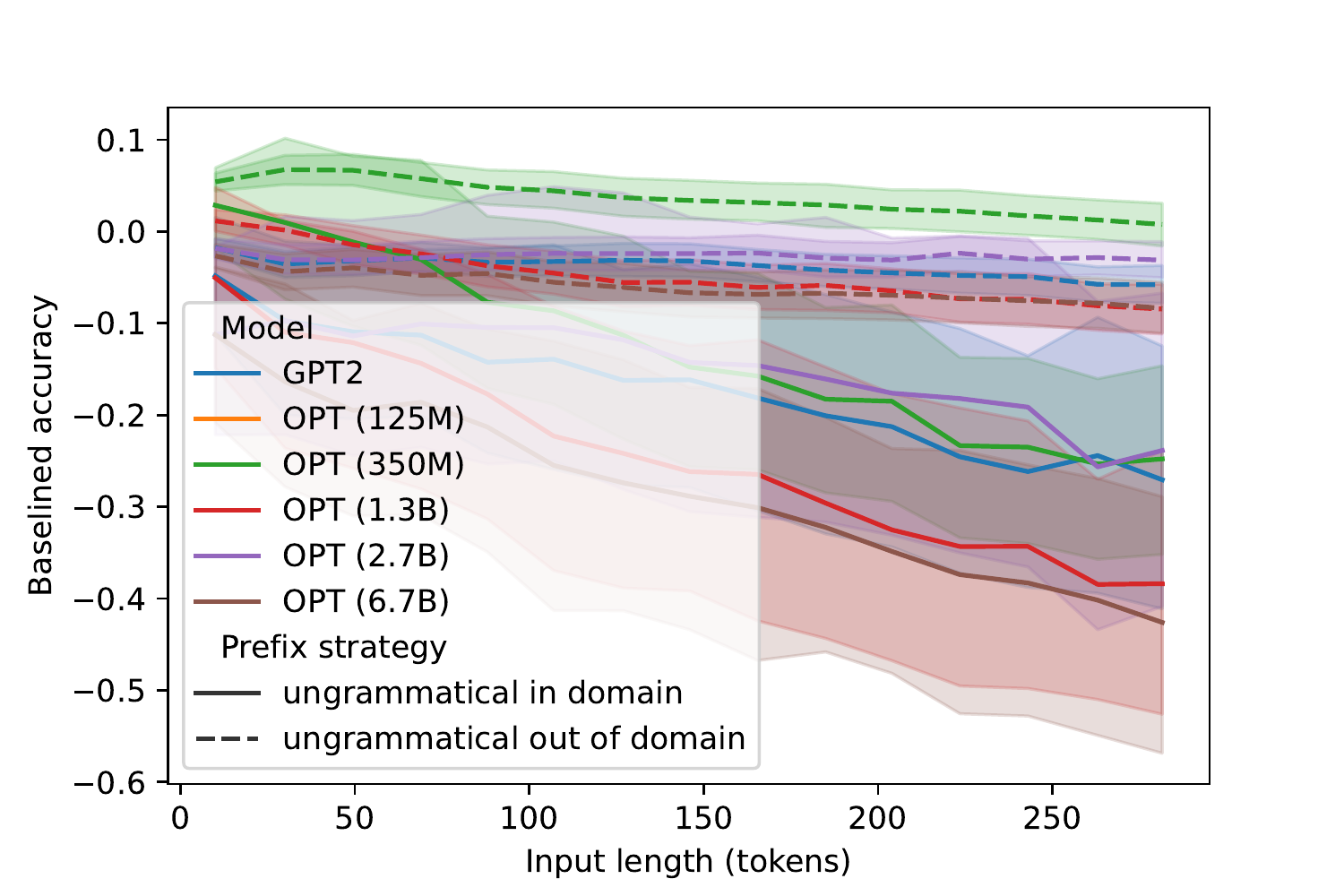}
\caption{Effects of length and domain per model for ungrammatical prefixes. Analogous to \autoref{fig:blimp_invsout_ungrammatical} in the main text.}
\label{fig:syntaxgym_invsout_ungrammatical}
\end{figure}

Figures \ref{fig:syntaxgym-positive-grammatical}, \ref{fig:syntaxgym_invsout_grammatical} and \ref{fig:syntaxgym_invsout_ungrammatical} are corresponding figures for the BLiMP results in Section \ref{sec:results}.

\end{document}